\documentclass[10pt,twocolumn,letterpaper]{article}

\usepackage{wacv}              

\usepackage{graphicx}
\usepackage{amsmath}
\usepackage{amssymb}
\usepackage{booktabs}
\usepackage[accsupp]{axessibility}

\usepackage[pagebackref,breaklinks,colorlinks]{hyperref}

\usepackage[capitalize]{cleveref}
\crefname{section}{Sec.}{Secs.}
\Crefname{section}{Section}{Sections}
\Crefname{table}{Table}{Tables}
\crefname{table}{Tab.}{Tabs.}

\usepackage{amsfonts}
\usepackage{multirow}
\usepackage{wrapfig}
\usepackage[framemethod=TikZ]{mdframed}
\usepackage{enumitem}

\usepackage{xcolor}

\NewDocumentCommand{\ying}{ mO{} }{\textcolor{teal}{\textsuperscript{\textit{Ying}}\textsf{\textbf{\small[#1]}}}}

\newcommand{\hide}[1]{}

\newcommand{\mpage}[2]
{
\begin{minipage}{#1\linewidth}\centering
#2
\end{minipage}
}

\usepackage{soul}
\usepackage{algorithm}
\usepackage{algpseudocode}
\usepackage{caption}
\usepackage{subcaption}
\usepackage{arydshln}

\DeclareMathOperator*{\argmax}{arg\,max}

\newcommand{\elbamoji}[0]{\smash{\raisebox{-2pt}{\includegraphics[height=0.30in]{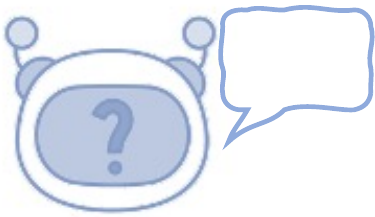}}}}

\newcommand*{\elbafont}{\fontfamily{augie}\selectfont}
\newcommand*\modelname{{\elbafont ELBA}\xspace}
\newcommand*\longmodelname{{\elbafont{E}}mbodied {\elbafont{L}}earning-{\elbafont{B}}y-{\elbafont{A}}sking\xspace}

\def\wacvPaperID{787} 
\def\confName{WACV}
\def\confYear{2025}

\begin{document}

\title{\elbamoji \modelname: Learning by Asking for Embodied Visual Navigation and Task Completion}

\author{
Ying Shen\\
University of Illinois Urbana - Champaign\\
{\tt\small ying22@illinois.edu}
\and
Daniel Biś\\
Amazon\\
{\tt\small bisdb@amazon.com}
\and
Cynthia Lu\\
Amazon\\
{\tt\small cynthilu@amazon.com}
\and
Ismini Lourentzou\\
University of Illinois Urbana - Champaign\\
{\tt\small lourent2@illinois.edu}
}
\maketitle


\begin{abstract}
The research community has shown increasing interest in designing intelligent embodied agents that can assist humans in accomplishing tasks. Although there have been significant advancements in related vision-language benchmarks, most prior work has focused on building agents that follow instructions rather than endowing agents the ability to ask questions to actively resolve ambiguities arising naturally in embodied environments. To address this gap, we propose an \longmodelname (\modelname) model that learns when and what questions to ask to dynamically acquire additional information for completing the task. We evaluate \modelname on the TEACh vision-dialog navigation and task completion dataset. Experimental results show that the proposed method achieves improved task performance compared to baseline models without question-answering capabilities. Code is available at \url{https://github.com/PLAN-Lab/ELBA}. 
\end{abstract}

  \section{Introduction}
The ultimate goal of embodied AI is to create interactive intelligent agents capable of assisting humans in various tasks. 
To achieve this, embodied agents must be able to understand instructions, interact seamlessly with humans, and resolve ambiguities arising in real-world scenarios. 
Over the past years, the research community has increasingly focused on designing embodied agents that can complete complex tasks through navigation and interaction with the environment. This has led to the emergence of a wide range of embodied AI tasks, including vision-language navigation \cite{mattersim, fried2018speaker}, vision-language task completion \cite{ALFRED20, padmakumar2021teach}, rearrangement \cite{batra2020rearrangement, weihs2021visual}, and embodied Question Answering (QA) \cite{das2018embodied, eqa_matterport}. These tasks address the challenges of endowing agents with various abilities, such as navigating to a specific location and manipulating multiple different objects within the environment. 

Despite recent advances in vision-language navigation and task completion, most prior work has focused on building agents that follow instructions \cite{thomason2020vision, wang2020environment, zhu2020vision, zhang2020diagnosing}. Current models often rely on human-curated training for supervision, without the ability to actively interact with the environment, which includes deciding when to acquire more information and determining which questions to ask in real-time when performing everyday tasks. 

\begin{figure}[t]
\centering
\includegraphics[width=0.8\linewidth]{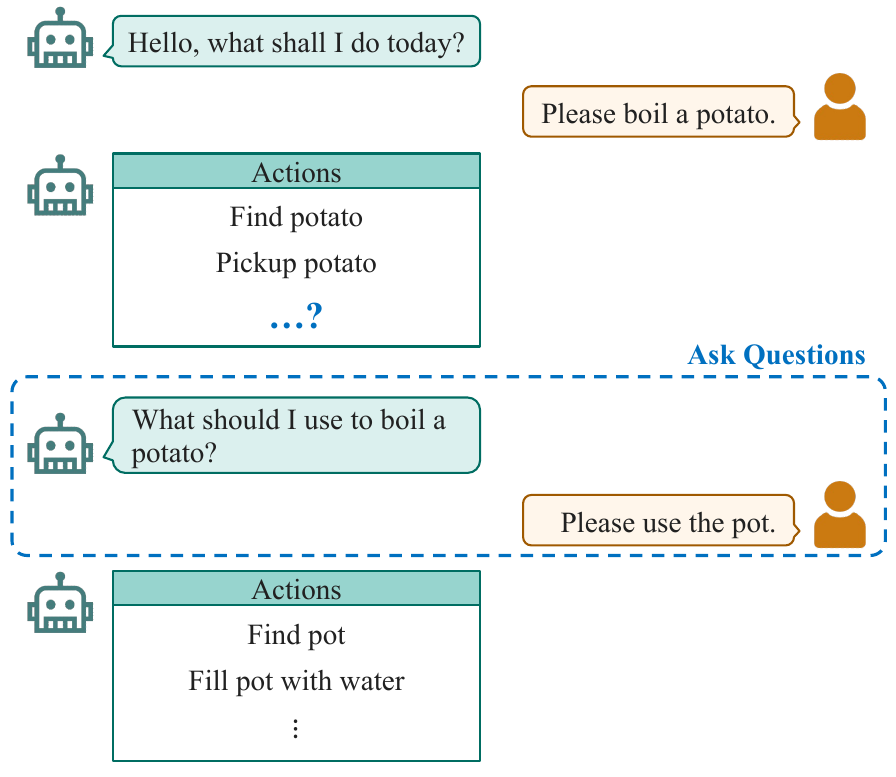}
\caption{\textbf{\longmodelname.} An example task \texttt{``Boil Potato''} which involves an agent (left) and an oracle (right). The goal for the agent is to complete the task by navigating and interacting with the environment. When uncertain about the next action, the agent can ask questions to the oracle, receive guidance, and proceed with more confidence to accomplish the task.}
\label{fig:teaser}
\end{figure}

A few attempts investigate the use of oracle answers for training interactive agents, avoiding the need for extensive human involvement~\cite{nguyen2019help, chi2020just,zhu2021self,shrivastava2021visitron,gao2022dialfred}. 
Yet, most methods are predominantly tailored to navigation tasks and are often constrained by the types and forms of oracle answers. 
For example, many rely on template-based approaches, which limits their applicability in open domains and hinders their capability to handle complex tasks and natural interactions. While conventional template-driven solutions excel in pure navigation scenarios, where agents can readily formulate questions and obtain well-defined ground truth answers, \eg, by calculating distances between current and target location, they fall short in vision-dialog task completion and object interaction tasks. In such settings, the best next step is often ambiguous, as there exist multiple valid ways to complete a task and no universally correct oracle-defined action is available for each timestep.

In this work, we introduce \longmodelname (\modelname), a model that learns \textit{when} to ask and  \textit{what} to ask when performing household tasks (\eg, Figure~\ref{fig:teaser}).
Currently, there is limited research on learning to ask questions in vision-dialog navigation and task completion, making our work distinct and one of the first works to tackle task-driven interactive embodied free-form QA.
\modelname integrates a newly introduced confusion module within the \textsc{Actioner}. At each timestep, the \textsc{Actioner} predicts the next action and object, while the confusion module evaluates the agent's uncertainty. When the confusion level is high, the agent is prompted to ask questions. We measure the confusion level using two approaches, entropy-based confusion, and gradient-based confusion. To generate helpful answers, \modelname includes a \textsc{Planner} module that predicts high-level future sub-goal instructions. Then, a \textsc{QA Generator} produces a set of free-form and template-based question-answer pairs, while a \textsc{QA Evaluator} selects the most relevant question based on a proposed contrastive relevance scoring method.

We evaluate \modelname on vision-language navigation and task completion and conduct ablation studies to analyze the impact of different confusion estimation methods and QA types on model performance.
Experimental results show \modelname achieves competitive performance compared to baselines, demonstrating the advantage of the proposed agent's ability to dynamically ask questions. Moreover, we verify that \modelname asks meaningful questions in a sample-efficient manner, reducing the number of questions per task by 57\% compared to a baseline that asks questions at fixed intervals.
In summary, our contributions are:
\textbf{(1)} We introduce an \longmodelname (\modelname) model that learns when and what questions to ask for vision-dialog navigation and task completion. In contrast to prior work, \modelname supports both template and free-form formats. \textbf{(2)} We demonstrate the effectiveness of the proposed approach and show that \modelname outperforms baselines. \textbf{(3)} We verify that the ability to dynamically ask questions improves task performance in embodied household tasks.

  \section{Related Work}

\noindent \textbf{Embodied Vision-Language Planning.}
Recent advancements in areas of embodied AI and multimodal machine learning have led to the emergence of various embodied vision-language planning tasks \cite{francis2022core}, such as Vision-Language Navigation (VLN) \cite{anderson2018vision, fried2018speaker,pashevich2021episodic,thomason2020vision,gao2023adaptive}, 
and Vision-Dialog Navigation and Task Completion \cite{padmakumar2021teach, narayan2019collaborative, suhr2019executing}. This family of works primarily focused on embodied navigation and object manipulation problems. Each task concentrated on distinct challenges, including navigation, object interaction, instruction following, and human-robot conversation.
Despite recent progress on these embodied vision-language planning tasks, most works have focused on building agents that understand instructions in the form of natural language \cite{anderson2018vision,fried2018speaker,pashevich2021episodic,gao2023adaptive} or simply use dialog as historical information alone \cite{de2018talk,thomason2020vision,wang2020environment,zhu2020vision} rather than endowing agents the ability to ask questions and actively acquire additional information. However, in practice, robots operating in human spaces need not only to understand and execute instructions but also to interact and resolve ambiguities arising naturally when performing complex tasks. Our work presents one of the first embodied agents that can dynamically decide when and what to ask for vision-dialog task completion.

\noindent \textbf{Multimodal Transformers.}
Transformer~\cite{vaswani2017attention} models have achieved remarkable success in natural language processing and have been effectively adapted in multimodal settings~\cite{team2023gemini,liu2024visual,xu-etal-2023-multiinstruct,huangembodied,wahed2024fine,zhang2024mm}, demonstrating improvements on various multimodal tasks, including visual question answering \cite{antol2015vqa} and vision-language navigation \cite{anderson2018vision}. 
Embodied vision-language planning tasks are inherently multimodal and require jointly learning representations of multiple modalities such as instructions, the sequence of observations along the corresponding trajectory, and the sequence of actions. This leads to the use of multimodal Transformer architectures designed explicitly for embodied vision-language planning \cite{hao2020towards,chen2021history,pashevich2021episodic,huangembodied}. 
The Episodic Transformer (E.T.)~\cite{pashevich2021episodic} utilizes separate encoders for language, vision, and action to encode modality-specific information and proposes a lightweight multimodal Transformer for vision-language planning.
Building upon the success of multimodal Transformer models in embodied vision-language planning tasks, our work aims to build an embodied agent that can learn when and what questions to ask during task performance.

\noindent \textbf{Visual Question Generation.}
Visual Question Generation (VQG) aims to generate questions from referenced visual content \cite{mostafazadeh2016generating,zhao2017video}. 
With the progress in embodied AI, prior works proposed visual question-answering tasks in embodied environments \cite{das2018embodied, gordon2018iqa}. 
However, these focus more on the agent's ability to plan actions in the environment in order to answer questions. 
Env-QA~\cite{gao2021env} focuses on evaluating the visual ability of environment understanding by asking the agent to answer questions based on an egocentric video composed of a series of actions that happened in the environment.
Our work draws inspiration from relevant VQG works and introduces an answer-aware question generator, enabling the embodied agent to ask task-relevant questions.

\noindent \textbf{Learning by Asking Questions.}
Recent works also learn to accomplish navigation tasks by 
determining when to ask questions \cite{chi2020just, nguyen2019help, zhu2021self, shrivastava2021visitron} and what to ask \cite{roman2020rmm,zhu2021self,sarch2023open}. However, they mainly focus on navigation tasks where ground truth answers are well-defined and thus it is relatively easy for the agent to ask questions to the oracle. 
DialFRED~\cite{gao2022dialfred} presents an embodied instruction-following benchmark, enabling agents to actively pose questions (from predefined question types) and leverage the obtained information to enhance the completion of household tasks. 
However, in all the aforementioned works, agents are confined to asking template-based questions and/or receiving template-based answers, limiting the diversity of question-answer pairs.
In contrast, our work distinguishes itself by empowering agents to ask questions in free form. 
Ask4Help~\cite{singh2022ask4help} proposes a policy enabling agents to request expert assistance as needed. Nonetheless, this work assumes the availability of an expert to provide oracle answers at any time step, which can be a costly resource, and such expert guidance may not always be accessible. 
In contrast, our method addresses the challenge of generating diverse and free-form question-and-answer pairs, without relying on the presence of an expert.

  \begin{figure*}[t!]
\centering
\includegraphics[width=0.95\linewidth]{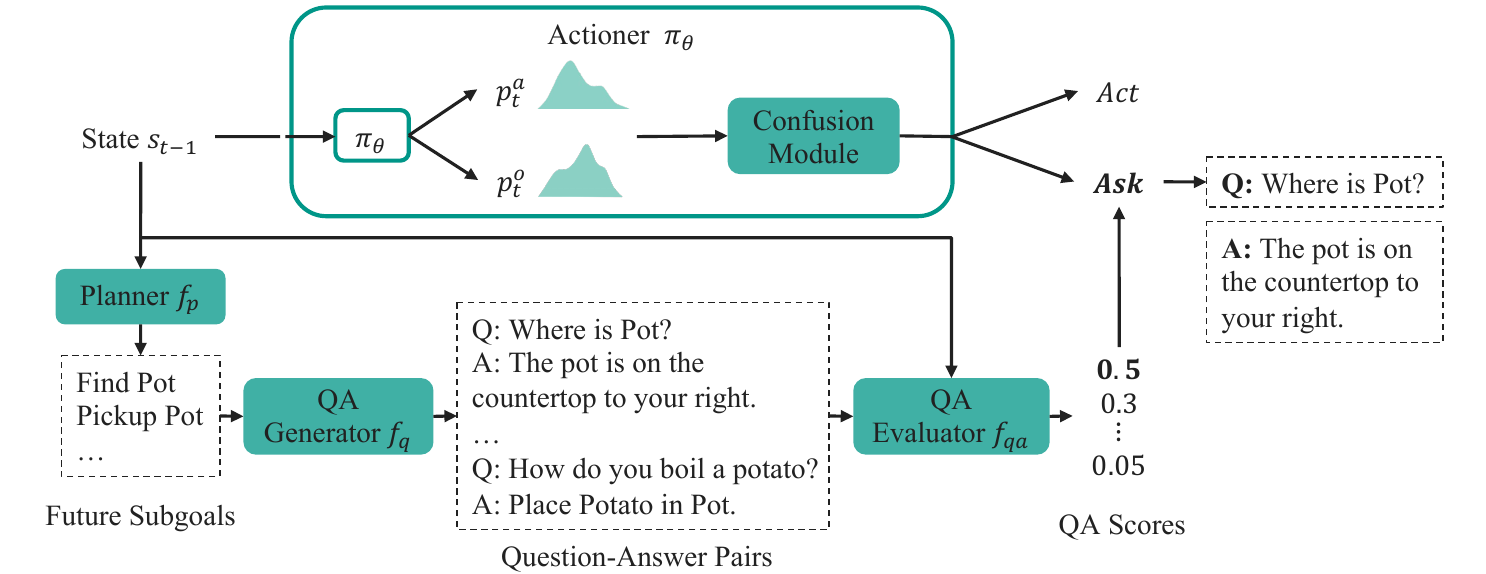}
\vspace{-0.2cm}
\caption{\textbf{\longmodelname (\modelname).} At every time step $t$, the \textsc{Actioner} encodes the state information $s_{t-1}$ and outputs the action and object distribution $(p_t^\alpha, p_t^o)$. The confusion module then determines the agent's confusion level by measuring either the entropy of the predicted distribution or the model gradient magnitude. If the confusion level exceeds a certain threshold, the agent will try to ask a question. Based on the state history, the \textsc{Planner} predicts high-level future sub-goal instructions which are later used to generate candidate answers. The \textsc{QA generator} then creates a set of candidate question-answer pairs based on the \textsc{Planner} outputs. The \textsc{QA evaluator} assigns a score to each QA pair, indicating their suitability for the current state, and ranks all QA pairs. The agent samples a pair from the top-$k$ QA pairs and asks the corresponding question if the confusion level decreases after incorporating the chosen QA pair.
}
\label{fig:lba}
\end{figure*}
\section{Problem Statement}
\label{sec:problem_stat}
Vision-Dialog Navigation and Task Completion requires an agent to engage in dialog, navigate, interact with the environment, and follow instructions to complete various tasks.
Each task trajectory is a tuple $(x_{1:T}, v_{1:T}, \alpha_{1:T})$ of natural language dialog utterances, visual observations, and physical actions, where $T$ is the trajectory length. The visual observations are a sequence of $T$ egocentric agent observations, \ie, $v_{1:T} \,{=}\, [v_1, \ldots, v_t, \ldots, v_T]$. The physical actions are a sequence of $T$ actions taken by the agent, \ie, $\alpha_{1:T} \,{=}\, [\alpha_1, \ldots, \alpha_t, \ldots, \alpha_T]$, where $\alpha_t \in \mathcal{A}$. The action space $\mathcal{A}\,{=}\,\{\mathcal{A}^N, \mathcal{A}^I\}$ consists of two types of actions: (1) navigation actions $\mathcal{A}^N$ that move the agent in discrete steps (\eg, \texttt{Turn Right}, \texttt{Pan Left}, \texttt{Forward}, etc.) and (2) interaction actions $\mathcal{A}^I$ that allow the agent to interact with the objects in the environment (\eg, \texttt{Pickup}, \texttt{Slice}, \texttt{Open}, \etc). The action and object distributions of the learned agent policy are denoted as $\pi_{\theta}(s_{t-1}) \,{=}\, (p_t^\alpha, p_t^o)$.

At each time step $t$, given the state information $s_{t-1} \,{=}\, (x_{1:t-1}, v_{1:t-1}, \alpha_{1:t-1})$, the agent must select the next action $\alpha_t \in \mathcal{A}$ to complete the task. To empower the agent with the ability to ask questions in ambiguous situations, we propose to learn when and what questions to ask in order to acquire additional information for completing the task. Thus, at each time step $t$, and before selecting the next action, the agent has to decide whether to form a question $q_t$ and include the respective question and answer pair $(q_t, a_t)$ in the state information. Then, given the augmented state information $\tilde{s}_{t-1} \,{=}\, (x_{1:t-1}, q_t, a_t, v_{1:t-1}, \alpha_{1:t-1})$, the agent will select the next action based on $\tilde{s}_{t-1}$. We describe the overall model architecture in the next section.

\section{\longmodelname}
\label{sec:approach}

The proposed \longmodelname model (\modelname) consists of four major components: an \textsc{Actioner}, a \textsc{Planner}, a \textsc{QA Generator}, and a \textsc{QA Evaluator}. 
At each time step $t$, the \textsc{Actioner} encodes the state information $s_{t-1}$ and predicts the next action and object distribution $\pi_{\theta}(s_{t-1}) \,{=}\, (p_t^\alpha, p_t^o)$. The  \textsc{Actioner}'s \textit{confusion module} then determines the agent's confusion level by measuring either the entropy of the predicted distribution or the gradient magnitude of the model (detailed in Subsection \ref{sec:actioner}). If the confusion level exceeds a threshold, the agent will attempt to ask a question. 
To first generate meaningful candidate answers, we introduce a \textsc{Planner} module that predicts high-level future sub-goal instructions. Then, the \textsc{QA Generator} constructs a set of $K$ candidate question-answer pairs $\mathcal{Q}_t \,{=}\, \{(q_t^i, a_t^i)\}_{i=1}^{K}$ by generating answer-aware questions,  informed by the \textsc{Planner} predicted sub-goals.
The \textsc{QA Evaluator} computes a relevance score $\phi(q_t^i, a_t^i), \forall i \,{=}\,1,\dots,K$ for each question-answer pair, evaluating how suitable each question-answer pair is for the current state, and then ranks and selects the top-$k$ pairs, \ie 
\begin{equation}
\mathcal{R}_t^* = \argmax\limits_{\mathcal{R}_t \subset \mathcal{Q}_t,|\mathcal{R}_t|=k} \sum\limits_{(q_t^i, a_t^i) \in \mathcal{R}_t} \phi(q_t^i, a_t^i).
\end{equation}
The agent samples a question and its corresponding answer $(q_t^{*}, a_t^{*}) \sim \mathcal{R}_t^*$ according to the normalized score distribution.
If the agent's confusion level decreases by incorporating the sampled question-answer pair into the state information, then the agent will ``ask'' the respective question before selecting its next action. Figure~\ref{fig:lba} presents an overview. 

\subsection{\textbf{\textsc{Actioner}}}\label{sec:actioner}
We build the \textsc{Actioner} based on the Episodic Transformer (E.T.) model \cite{pashevich2021episodic}, a multimodal Transformer \cite{vaswani2017attention} that encodes state information, including visual observations, actions, and dialog, and then predicts the following action and the possible object involved. 

\noindent \textbf{Encoder:} At each time step $t$, the \textsc{Actioner} encodes the state information $s_{t-1} \,{=}\, (x_{1:t-1}, v_{1:t-1}, \alpha_{1:t-1})$ including the history utterances $x_{1:t-1}$, visual observations $v_{1:t-1}$, and physical actions $\alpha_{1:t-1}$ via a multimodal encoder ${f}_{e}(\cdot)$:
\begin{equation}\label{state_encoder}
    h_{t-1} = {f}_{e}(s_{t-1}),
\end{equation}
where $h_{t-1}$ refers to the multimodal hidden state.

\noindent \textbf{Decoder:} 
The encoded hidden state $h_{t-1}$ is passed through two multilayer perceptrons $\text{MLP}_\alpha(\cdot)$ and $\text{MLP}_o(\cdot)$ to predict the probability vector $p^\alpha_t$ and $p^o_t$ of the next action and corresponding object, respectively:
\begin{align}
    p^\alpha_t = f_\alpha(h_{t-1}) = \sigma(\text{MLP}_\alpha(h_{t-1})) \\
    p^o_t = f_o(h_{t-1}) =\sigma(\text{MLP}_o(h_{t-1})),
\end{align}
where $\sigma(\cdot)$ is the softmax activation, $f_\alpha$ and $f_o$ are the decoder networks for action and object.

\noindent \textbf{Confusion Module:}
The \textsc{Actioner} can only predict navigation and interaction actions and is not capable of asking questions. We introduce a confusion module that allows the agent to decide \textit{when} to ask questions based on its confusion levels. We formalize agent confusion in two different ways: entropy-based or gradient-based confusion.

\noindent For the \textbf{entropy-based confusion}, we model the confusion level through the entropy of the predicted action and object probability distributions, $p^\alpha_t$ and $p^o_t$. 
At each time step $t$, the agent will try to generate a question if the entropy of the action distribution is greater than a threshold  $H(p_t^\alpha) > \epsilon_\alpha$ or if the entropy of the object distribution is greater than a threshold and the predicted action is an interaction action, \ie, $H(p_t^o) > \epsilon_o$ and $\hat{\alpha}_t \in \mathcal{A}^{I}$. The thresholds $\epsilon_\alpha, \epsilon_o$, therefore, control the overall confusion level. 
Finally, if the entropy is decreased by incorporating the sampled question-answer pair in the augmented state information $\tilde{s}_{t-1} \,{=}\, (x_{1:t-1}, q_t^{*}, a_t^{*}, v_{1:t-1}, \alpha_{1:t-1})$, the agent will ask the generated question. 

\noindent On the other hand, \textbf{gradient-based confusion} models the confusion level through the gradient magnitude. We measure the agent's confusion level by computing the gradient $g_t$ of the loss w.r.t. the multimodal hidden state $h_{t-1}$.
\begin{equation}
    g_t = \nabla_{h_{t-1}} \Big(\mathcal{L}\left(f_\alpha(h_{t-1}), \alpha_t\right) + \mathcal{L}\left(f_o(h_{t-1}), o_t\right)\Big),
\end{equation}
where $\mathcal{L}(\cdot, \cdot)$ denotes the loss function, $f_\alpha(\cdot, \cdot)$ and $f_o(\cdot, \cdot)$ are the decoder networks for action and object, respectively.
However, a challenge in computing this gradient is that it requires knowledge of the ground truth action and object $(\alpha_t^*, o_t^*)$. In batch active learning settings, BADGE~\cite{ash2019deep} proposes to treat the model's prediction as the ground truth pseudo-label and proves that the gradient norm using this pseudo-label serves as the lower bound of the gradient norm induced by the ground-truth label.
Specifically, given the predicted action and object probability distributions $(p^\alpha_t, p^o_t)$ at time step $t$, the most likely action $\hat{\alpha_t}$ and object $\hat{o_t}$ can be formalized as
\begin{equation}
    \hat{\alpha_t} = \argmax\limits_{\alpha} p_t^\alpha, \;\;\;\; \hat{o_t} = \argmax\limits_{o} p_t^o.
\end{equation}

Therefore, by replacing the ground truth action and object $(\alpha^*_t, o^*_t)$ with the predicted $(\hat{\alpha_t}, \hat{o_t})$, we can compute the gradient of the hidden state $h_{t-1}$ as:
\begin{equation}
    g_t = \nabla_{h_{t-1}} \Big(\mathcal{L}\left(f_\alpha(h_{t-1}), \hat{\alpha_t}\right) + \mathcal{L}\left(f_o(h_{t-1}), \hat{o_t}\right)\Big).
\end{equation}

Finally, we use the $\ell_2$ norm of the gradient $\|g_t\|_2$ as the measure of confusion level, where a large norm indicates high uncertainty in the model's prediction  \cite{ash2019deep}. Similar to the entropy-based approach, the agent will try to generate a question if the norm of the gradient is greater than a threshold $\|g_t\|_2 > \epsilon$, and the agent will ask the selected question if it results in a decrease in confusion.

\subsection{\textbf{\textsc{Planner}}}
In contrast to navigation tasks, where the optimal next step is well-defined and often used as an oracle answer, determining the best next step in more complex tasks can be challenging. To address this, we employ sub-goal instructions as potential answers, as they provide information about the possible optimal next step. Specifically, we propose a \textsc{Planner} module to generate high-level future sub-goal instructions. Sub-goals can be viewed as high-level sub-tasks of a particular task. For example, given the task \texttt{``Make coffee''}, the sequence of sub-goals could be \texttt{``Find coffee machine''} $\rightarrow$ \texttt{``Find mug''} $\rightarrow$ \texttt{``Pickup mug''}, \etc These sub-goals can thus be used as candidate answers for generating answer-aware questions.
At each time step $t$, the \textsc{Planner} $f_p(\cdot)$ takes the concatenation of natural language utterances $x_{1:t-1}$ and all possible actions $\alpha_{1:|\mathcal{A}|} = [\alpha_1, \alpha_2, \cdots, \alpha_{|\mathcal{A}|}]$ as input and generates a future sub-goal sequence
\begin{equation}
    z_{t:T} = f_p([x_{1:t-1};\alpha_{1:|\mathcal{A}|}]),
\end{equation}
where $[;]$ denotes the concatenation operation and $\mathcal{A}$ is the space of all physical actions from the pre-defined action set, \ie, $\mathcal{A}_I \cup \mathcal{A}_N$. We employ a pre-trained T5 model \cite{raffel2019exploring}, fine-tuned on the training dataset, as the backbone of the encoder-decoder in \textsc{Planner}.
 
\subsection{\textbf{\textsc{QA Generator}}}
The \textsc{QA Generator} generates a set of candidate question-answer pairs, which include two types, oracle (template-based) and model-generated (free-form) pairs.

\noindent \textbf{Oracle QA Pairs:}
Oracle QA pairs are generated using five types of pre-defined question-answer templates: \texttt{Location}, \texttt{Appearance}, \texttt{Current/Next Sub-goals}, 
and \texttt{Direction}. Table \ref{table:oracle_qa} shows examples of the templates for generating oracle QA pairs. We define the first three template types similar to \cite{gao2022dialfred} 
by utilizing available object attribute information such as object material, and the agent's location in the simulated environment. For the \texttt{Current/Next Sub-goals} template, we leverage the outputs from the \textsc{Planner}.

\begin{table}[!t]
  \centering
    \caption{\textbf{Oracle QA Templates.}}
    \vspace{-0.3cm}
  \resizebox{\linewidth}{!}{%
\begin{tabular}{ll}
\toprule
\textbf{QA Type} & \textbf{QA Template} \\ 
\midrule
\multirow{2}{*}{Location} & \textbf{Q:} Where is [\textit{object}]? \\
 & \textbf{A:} The [\textit{object}] is in/on the [\textit{container}] [\textit{direction}]. \\ \hdashline
\multirow{2}{*}{Appearance} & \textbf{Q:} What does [\textit{object}] look like?\\
    & \textbf{A:} The [\textit{object}] is made of [\textit{material}]. \\ \hdashline
\multirow{2}{*}{Direction} &  \textbf{Q:} Which [\textit{direction}] should I turn to?\\
  &  \textbf{A:} You should turn [\textit{direction}] / You don’t need to move. \\ \hdashline
\multirow{2}{*}{Current/Next Sub-goals} &  \textbf{Q:} What is current/next sub-goal?\\
  &  \textbf{A:} [\textit{current/next sub-goal}]. \\
 \bottomrule
\end{tabular}%
}
  \label{table:oracle_qa}
\end{table}

\noindent \textbf{Model-Generated QA Pairs:}
For the model-generated QA pairs, we first extract a set of candidate answers from generated future sub-goals $z_{t:T}$. Specifically, we first parse and extract all nouns from future sub-goals $z_{t:T}$. Then, we construct the set of candidate answers $\{a^i_t\}_{i=1}^K$ using the sub-goals $z_{t:T}$ and all the nouns extracted from each sub-goal. 
For example, given the sub-goal \texttt{``Pickup potato''}, the constructed candidate answers are \texttt{$\{\text{potato}, \text{ pickup potato}\}$}, and given the next sub-goal  \texttt{``Place the potato on the desk''}, the constructed candidate answers are \texttt{$\{\text{potato}, \text{ desk}, \text{ place potato on desk}\}$}. After removing repeated answers, the candidate answer set becomes \texttt{$\{\text{potato}, \text{ pickup  potato}, \text{ desk}, \text{ place} \\ \text{potato on desk}\}$}. 
Finally, given the history utterance $x_{1:t-1}$ and each extracted answer $a_t^i$, we generate an answer-aware question using a Transformer model $f_q(\cdot, \cdot)$:
\begin{equation}
    q^i_t = f_q(x_{1:t-1}, a^i_t).
\end{equation}
Specifically, we utilize a T5 model \cite{raffel2019exploring} finetuned on the SQuAD question generation dataset \cite{mromero2021t5-base-finetuned-question-generation-ap}.

\subsection{\textbf{\textsc{QA Evaluator}}}
The \textsc{QA Evaluator} assigns a relevance score $\phi(q^i_t, a^i_t)$ to each candidate question-answer pair by measuring the similarity between the state information and the question-answer pair. Based on the current state information, the most suitable question-answer pair should have the highest similarity score among all candidate pairs. Since the current state encompasses historical information (\ie, history of utterances, visual observations, and physical actions), the selected question-answer pair reflects not only the immediate contextual relevance but also integrates relevant historical contextual knowledge.
We finetune a DistilBERT \cite{sanh2019distilbert} model $f_{qa}(\cdot)$ to embed a question-answer pair $[q_t^i;a_t^i]$. 
Following BERT~\cite{kenton2019bert}, we adopt the hidden state corresponding to the reserved special \texttt{[CLS]} token as the embedding for the question-answer pair, denoted as $h_{qa_{i,t}}$. 
To embed state information, we use a multilayer perceptron projection layer and encode the hidden state information $h_{t-1}$ from the \textsc{Actioner}. We denote the embedding of the state information $s_t$ at time step $t$ as $h_{s_t}$. 
We measure the score $\phi(q^i_t, a^i_t)$ using the dot product between the $\ell_2$-normalized state information and question-answer pair embeddings,
\begin{equation}
    \phi(q^i_t, a^i_t) = \frac{h_{s_t}}{\|h_{s_t}\|} \cdot \frac{h_{qa_{i,t}}}{\|h_{qa_{i,t}}\|}.
\end{equation}

At training time, we sample a minibatch of $N$ pairs $\{(h_{s_t}, h_{qa_{i,t}})\}_{i=1}^{N}$ from training data, and the \textsc{QA Evaluator} is trained to maximize the similarity of the $N$ real pairs in the batch while minimizing the similarity of the embeddings of the $N^2 - N$ incorrect pairings. We adopt the CLIP $N$-pair contrastive loss~\cite{radford2021learning}.

  \section{Experiments}
\label{sec:exp}

\subsection{Experimental Setup}
\noindent \textbf{Dataset and Baselines.}
We train and evaluate \modelname on TEACh \cite{padmakumar2021teach}, a dataset of over $3,000$ human-human interactive dialogues to complete household tasks in the AI2-THOR simulation \cite{kolve2017ai2}. 
For evaluation, we experiment on the EDH instances using the divided test seen and unseen splits of TEACh\footnote{\url{https://github.com/alexa/teach\#teach-edh-offline-evaluation}}. The unseen test split consists of rooms that are unseen during training, while the seen test split contains rooms that are present/seen during training. 
We directly compare our proposed \modelname model with the E.T. baseline \cite{padmakumar2021teach}, which does not possess the ability to ask questions. The E.T. model baseline can be viewed as \modelname with only the \textsc{Actioner} module. Implementation details and hyperparameters can be found in the Appendix. 

\noindent \textbf{Evaluation Metrics.}
Following existing works \cite{padmakumar2021teach,ALFRED20}, we measure the task success and goal-condition success rates. 
\begin{itemize}[noitemsep, left=0pt, topsep=0pt]
\item \textbf{Goal-Condition Success Rate (GC):} Each task can contain multiple goal conditions, where successfully completing one single goal condition could require a lengthy sequence of actions. 
The goal-condition success rate is the ratio of goal conditions completed at the end of each episode, averaged across all episode trajectories. 

\item \textbf{Task Success Rate (SR):} Task success is defined as $1$ if all goal conditions have been completed at the end of the episode and $0$ otherwise. The final score is calculated as the average across all episodes.

\item \textbf{Trajectory Length Weighted (TLW) Metrics:} Path-weighted versions of both SR and GC metrics consider the length of the action sequence. For a reference trajectory $L$ and inferred trajectory $\hat L$, we calculate the Trajectory Length Weighted Task Success Rate (SR [TLW]), defined as
\begin{equation}
    \text{SR [TLW]} = \text{SR} * \frac{|L|}{\max(|L|, |\hat L|)}.
\end{equation}
The Trajectory Length Weighted Goal-Condition Success Rate (GC [TLW]) is computed similarly.
\end{itemize}

  \subsection{Quantitative Evaluation}
\label{sec:results}
We design experiments to answer the following research questions:
\textbf{(1) Impact of Asking Questions:} Direct comparison between our proposed \modelname model and the previous E.T. baseline model \cite{padmakumar2021teach} that does not possess the ability to ask questions.
\textbf{(2) Effect of Types of Questions:} We explore the performance of \modelname with different types of questions, oracle (template), model-generated (free-form), and a combination thereof.
\textbf{(3) Effect of Question Timing:} We evaluate \modelname variants with two different question timings, ask when confused and ask at fixed time steps.
\textbf{(4) Robustness of Confusion:} We evaluate the performance of \modelname with different confusion modules and thresholds.
\begin{table}[!t]
\centering
\caption{\textbf{Task and Goal-Condition Success.} Comparing E.T. baseline with \modelname variations with entropy-based confusion (\modelname w/E) and gradient-based confusion (\modelname w/G).
Trajectory length weighted metrics are presented in [brackets]. All values are percentages ($\%$). For all metrics, higher is better. Best performance is highlighted in \textbf{\textcolor{blue}{bold}}. We perform each experiment three times and report the average score.
}
\vspace{-0.2cm}
\resizebox{\columnwidth}{!}{%
\begin{tabular}{l c c c c}
\toprule
& \multicolumn{2}{c}{\textbf{Seen}} & \multicolumn{2}{c}{\textbf{Unseen}} \\
\textbf{Model} & \textbf{SR [TLW]} & \textbf{GC [TLW]} & \textbf{SR [TLW]} & \textbf{GC [TLW]} \\
\midrule
Baseline (E.T.) &  15.1{\scriptsize{$\pm$0.3}} [2.3{\scriptsize{$\pm$0.5}}] & 15.7{\scriptsize{$\pm$0.7}} [4.0{\scriptsize{$\pm$0.3}}] & 4.9 {\scriptsize{$\pm$ 0.1}} [0.2{\scriptsize{$\pm$0.0}}] & 3.3{\scriptsize{$\pm$0.1}} [0.8{\scriptsize{$\pm$0.0}}] \\ 
\modelname w/E & \textbf{\textcolor{blue}{15.8}}{\scriptsize{$\pm$0.2}} [1.6{\scriptsize{$\pm$0.4}}] & \textbf{\textcolor{blue}{19.2}}{\scriptsize{$\pm$0.8}} [4.1{\scriptsize{$\pm$0.4}}] & \textbf{\textcolor{blue}{5.7}}{\scriptsize{$\pm$0.0}} [0.5{\scriptsize{$\pm$0.0}}] & \textbf{\textcolor{blue}{3.8}}{\scriptsize{$\pm$0.0}} [1.1{\scriptsize{$\pm$0.0}}] \\
\modelname w/G & 15.4{\scriptsize{$\pm$0.2}} [1.8{\scriptsize{$\pm$ 0.1}}] & 18.4{\scriptsize{$\pm$1.9}} [3.9{\scriptsize{$\pm$0.5}}] & 5.1{\scriptsize{$\pm$0.1}} [0.2{\scriptsize{$\pm$0.0}}] & \textbf{\textcolor{blue}{3.8}}{\scriptsize{$\pm$0.0}} [1.1{\scriptsize{$\pm$0.1}}] \\
\bottomrule
\end{tabular}
}
\vspace{-0.3cm}
\label{table:sr_gc}
\end{table}

\noindent \textbf{Impact of Asking Questions:}
In Table \ref{table:sr_gc}, we report average model performance across three experimental trials. 
We observe that \modelname variants outperform the E.T. baseline on all metrics. Notably, \modelname w/E exhibits significant improvements in goal-condition success rate (GC), with relative performance gains of $22.29\%$ and $15.15\%$ on seen and unseen environments, respectively. Similarly, \modelname w/G showcases considerable enhancements, with a relative performance gain of $17.20\%$ and $15.15\%$ in goal-condition success rate (GC) for the seen and unseen environments, respectively.
As the E.T. model is a standalone \textsc{Actioner}, this competitive performance of \modelname compared to E.T. emphasizes the advantage of asking questions. Overall, \modelname with entropy-based confusion (\modelname w/E) achieves relatively better performance on both seen and unseen test splits as compared to the gradient-based confusion method (\modelname w/G).

Success rate, \ie, successful completion of all goal conditions within a single episode, requires numerous actions to be taken and several subgoals to be completed successfully. Considering the task complexity and the number of actions required to complete each goal condition, our results offer promising and encouraging insights. Both \modelname variants demonstrate their generalization ability by achieving improvements on the unseen test split, encompassing rooms that were not part of the training data. \modelname w/E exhibits significant improvements in unseen environments, with a relative performance gain of $15.15\%$ in goal-condition success rate (GC) and $16.33\%$ in success rate (SR). Similarly, \modelname w/G shows considerable improvements in unseen environments, with a relative performance gain of $15.15\%$ in GC and $4.08\%$ in SR. Apart from these performance improvements, our work is among the first to introduce open-ended questions in embodied environments beyond simple navigation, paving the way for future research in task-driven interactive embodied QA.

To further demonstrate the benefits of enabling an \textsc{ACTIONER} agent to ask questions when encountering confusion and improve performance through effective feedback, we extend our methodology to other types of \textsc{ACTIONERs}, specifically the HELPER~\cite{sarch2023open} model, by integrating a question-answering (QA) module. Our results, detailed in Appendix \ref{app:helper}, indicate that task-driven QA boosts HELPER's performance, suggesting that the proposed QA feedback mechanisms are potentially beneficial and adaptable across a variety of \textsc{ACTIONER} frameworks.

\noindent \textbf{Effect of Types of Questions:} 
We conduct ablation studies to investigate the effectiveness of different types of questions: (1) {Oracle QA}: template-based oracle questions, (2) {Generated QA}: free-form model-generated questions, (3) {Combined QA}: the combination of Oracle and {Generated QA}. 
In Table \ref{table:ablaQ}, we report the performance of \modelname w/E and \modelname w/G when the agent is only allowed to ask one type of question. 
The results show that both the model with template-based oracle QAs (\modelname ~--~ Oracle QA) and the model with free-form model-generated QAs (\modelname ~--~ Generated QA) outperform the baseline model. We also observe that either the variation with the model-generated QAs (\modelname ~--~ Generated QA) or the model that combines both oracle and model-generated QAs (\modelname ~--~ Combined QA) achieve the best performance on most metrics for the unseen test split. This indicates the effectiveness of combining template-based with free-form questions.
\begin{table}[!t]
\centering
\caption{\textbf{Ablation study on Question Types.} 
Trajectory length weighted metrics are presented in [ brackets ]. All values are percentages ($\%$). For all metrics, higher is better. Best performance is highlighted in \textbf{\textcolor{blue}{bold}}. Results averaged over two experimental runs.
}
\vspace{-0.2cm}
\resizebox{\columnwidth}{!}{%
\begin{tabular}{@{}p{1cm} p{4cm} p{2cm} p{2cm} p{2cm} p{2cm}@{}}
\toprule
& & \multicolumn{2}{c}{\textbf{Seen}} & \multicolumn{2}{c}{\textbf{Unseen}} \\
\cmidrule(lr){3-4} \cmidrule(lr){5-6}
& \textbf{Model} & \textbf{SR [TLW]} & \textbf{GC [TLW]} & \textbf{SR [TLW]} & \textbf{GC [TLW]} \\
\midrule
& Baseline (E.T.) &  15.1 [2.3] & 15.7 [4.0] & 4.9 [0.2] & 3.3 [0.8] \\ 
\midrule
\multirow{3}{*}{\rotatebox[origin=c]{90}{\parbox[c]{1cm}{\centering \textbf{\textcolor{gray}{\modelname w/E}}}}} & \modelname ~--~ Oracle QA &  \textbf{\textcolor{blue}{16.0}} [1.5] & \textbf{\textcolor{blue}{19.4}} [4.4] & 4.9 [0.2] & 3.6 [0.9]\\ 
& \modelname ~--~ Generated QA & 15.6 {[1.9]} & 19.0 [4.2] & 4.9 [0.2] & 3.6 [0.9]  \\ 
& \modelname ~--~ Combined QA & 15.8 [1.6] & 19.2 [4.1] & \textbf{\textcolor{blue}{5.7}} [0.5] & \textbf{\textcolor{blue}{3.8}} {[1.1]}\\
\midrule
\multirow{3}{*}{\rotatebox[origin=c]{90}{\parbox[c]{1cm}{\centering \textbf{\textcolor{gray}{\modelname w/G}}}}} & \modelname ~--~ Oracle QA & 15.4 [1.6] & 17.8 {[4.2]}  & 5.0 [0.2] & \textbf{\textcolor{blue}{3.8}} {[1.1]} \\ 
& \modelname ~--~ Generated QA & \textbf{\textcolor{blue}{15.5}} {[2.2]} & \textbf{\textcolor{blue}{18.6}} [4.1] & \textbf{\textcolor{blue}{5.4}} [0.3] & \textbf{\textcolor{blue}{3.8}} {[1.1]}\\ 
& \modelname ~--~ Combined QA & 15.4 [1.8] & 18.4 [3.9] & 5.1 [0.2] & \textbf{\textcolor{blue}{3.8}} [1.0]\\
\bottomrule
\end{tabular}
}
\label{table:ablaQ}
\vspace{-0.3cm}
\end{table}

\begin{figure}[t]
\centering
\includegraphics[width=0.99\columnwidth]{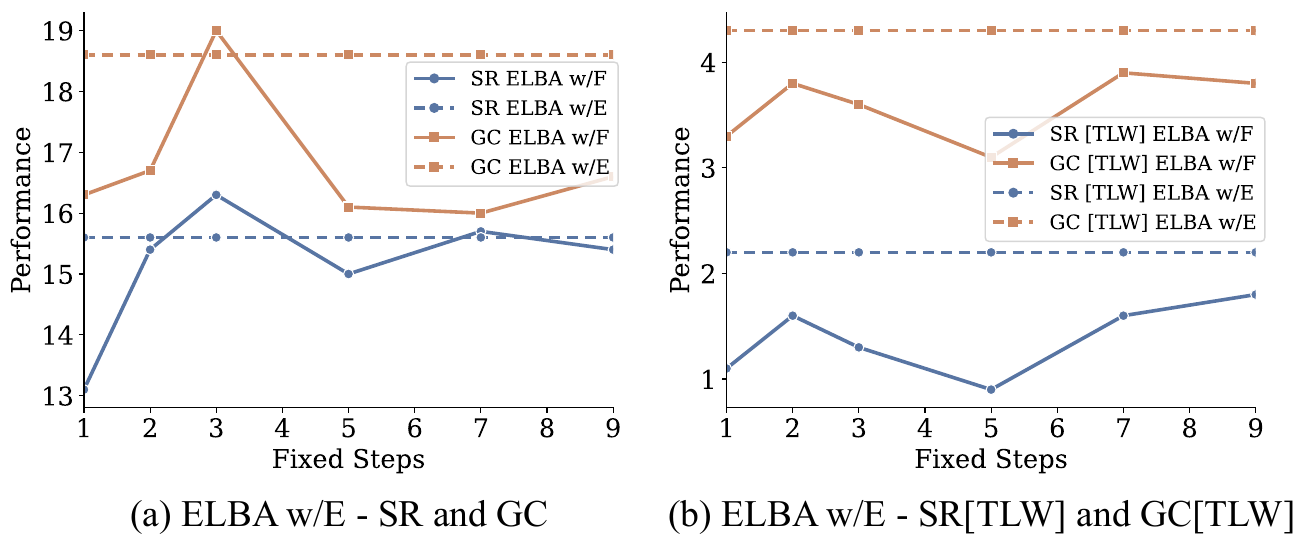}
\vspace{-0.3cm}
\caption{\textbf{Performance of \modelname ~--~ Oracle QA on question timing.} For \modelname w/F model variants, we control the number of fixed time steps the \textsc{Actioner} needs to execute before asking a question. Dashed lines show the performance of \modelname w/E with the proposed confusion module, while solid lines present \modelname w/F model variations with fixed time steps of asking questions.}
\label{fig:steps}
\end{figure}

\noindent \textbf{Effect of Question Timing:}
We also measure the change in performance with oracle QAs while varying the question timing, \ie, evaluating whether asking when confused produces better performance than asking at fixed time step intervals (we refer to such model variations as \modelname w/F). Figure \ref{fig:steps} reports the performance of \modelname ~--~ Oracle QA with two question timing variants on the test seen split, showing that, generally, the model that asks when confused outperforms the ones that ask at fixed time steps. We also observe that the model that asks every three time steps outperforms the model with the confusion module (\modelname w/E ~--~ Oracle QA). However, our model with the confusion module only requires asking $44 \pm 64$ (mean $\pm$ standard deviation) questions per task, while the model that asks every three steps needs to ask $102 \pm 101$ questions per task. This shows that the confusion module could help the agent perform reasonably well while asking fewer questions.

\begin{figure}[t]
\centering
\includegraphics[width=\columnwidth]{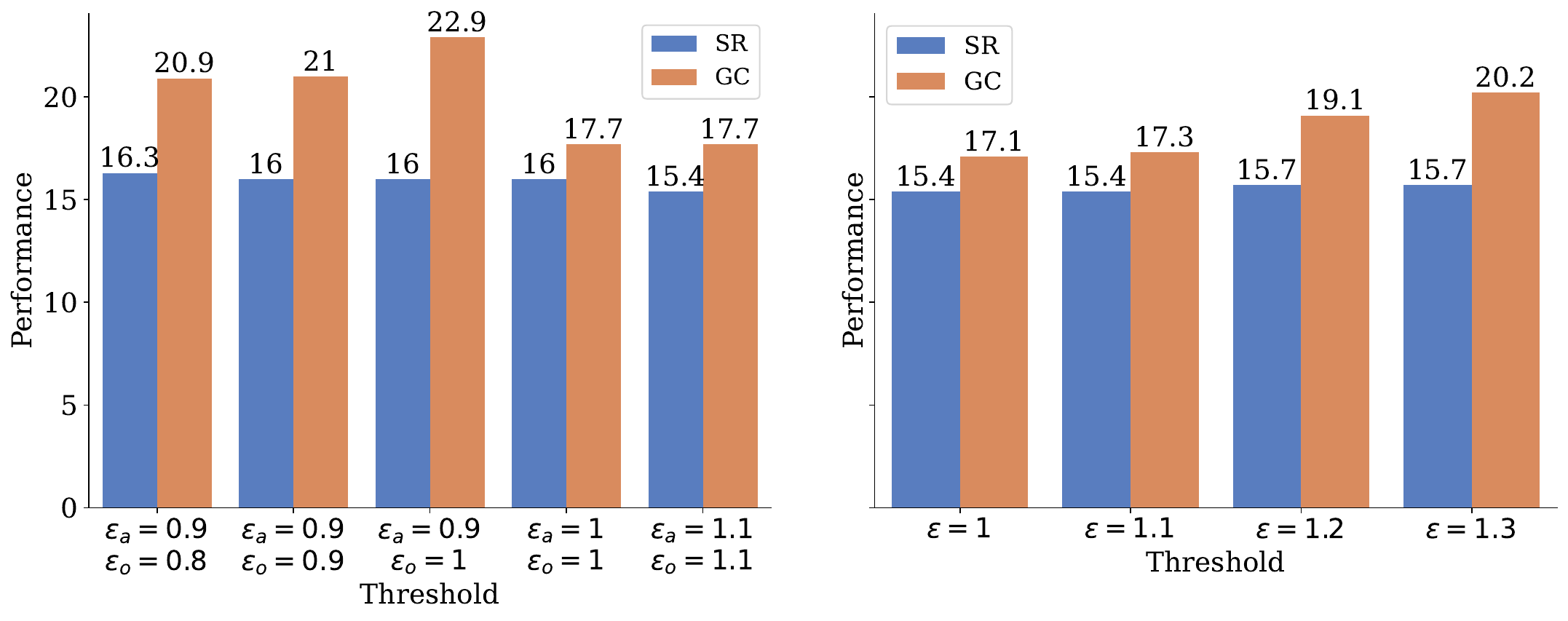}\hfill
\mpage{0.5}{(a) \modelname w/E}
\hfill
\mpage{0.45}{(b) \modelname w/G}
\vspace{-0.2cm}
\caption{\textbf{Varying confusion thresholds.} Performance of (a) entropy-based (\modelname w/E) and (b) gradient-based (\modelname w/G) using different thresholds for action and object distributions.}
\label{fig:threshold}
\vspace{-0.3cm}
\end{figure}

\noindent \textbf{Robustness of Confusion Threshold:}
\modelname allows the agent to ask questions when the confusion level is greater than a threshold. Therefore, different threshold settings could impact model performance. 
Here, we investigate the robustness of the confusion module against minor fluctuations around the optimally chosen threshold, which was determined using the validation split.
We first identify the optimal confusion threshold that yields the highest performance on the validation set and then examine how the model's performance fluctuates when we adjust the confusion thresholds around this optimal point. For entropy-based confusion, we measure the change in performance on TEACh while varying the thresholds $\epsilon_a$ and $\epsilon_o$ for action distribution and object distribution, respectively. Similarly, for gradient-based confusion, we measure the change in performance while varying the gradient norm threshold $\epsilon$.
In Figure \ref{fig:threshold}, we observe that both \modelname w/E and \modelname w/G outperform the baseline on the test seen split with different combinations of threshold settings, showing the robustness of threshold selection. We also report the performance of \modelname w/E with different thresholds for the action and object distributions and find that using a common threshold for both does not substantially affect performance.

\begin{figure}[t]
\centering
  \begin{subfigure}[b]{0.49\columnwidth}
    \includegraphics[width=\textwidth]{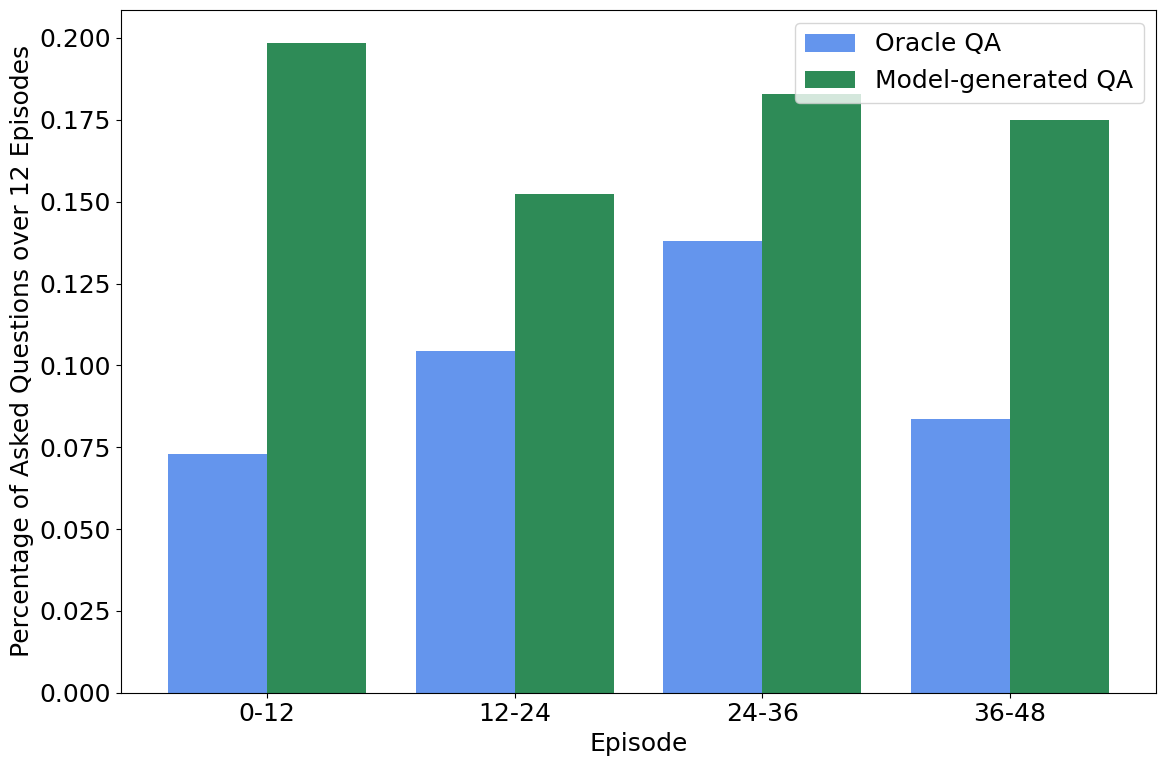}
    \caption{\modelname w/E}
    \label{fig:f1}
  \end{subfigure}
  \begin{subfigure}[b]{0.49\columnwidth}
    \includegraphics[width=\textwidth]{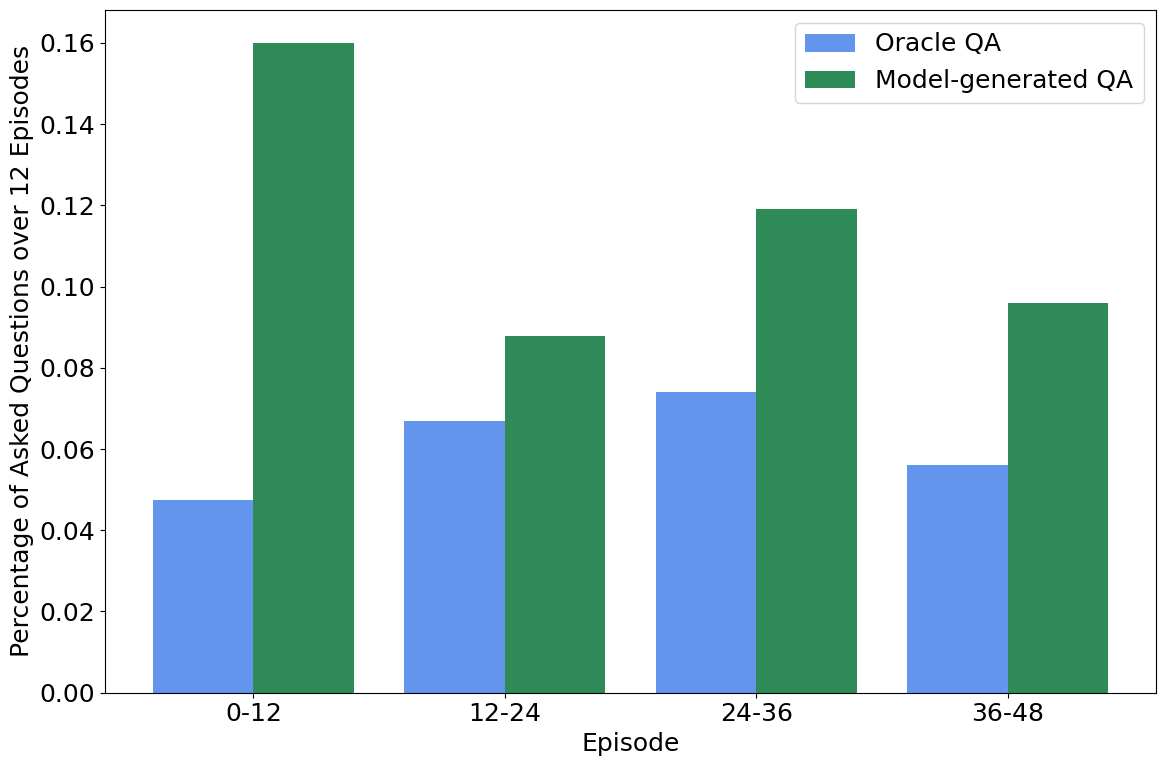}
    \caption{\modelname w/G}
    \label{fig:f2}
  \end{subfigure}
\vspace{-0.3cm}
\caption{\textbf{Average percentage of asked questions} for oracle QA and model-generated QA in successful episodes with (a) entropy-based (\modelname w/E) and (b) gradient-based (\modelname w/G).}
\label{fig:ques_types}
\end{figure}

\noindent \textbf{Distribution of Question Types:}
\modelname with Combined QA allows the agent to ask both oracle and model-generated questions. Figure \ref{fig:ques_types} shows the distribution of the different types of questions asked by the agent in successful episodes. We measure the average percentage of asked questions per type, normalized by the current episode trajectory length. We report the aggregated average percentage of asked questions over 12 episodes. We find that, for successful episodes, \modelname with Combined QA is more likely to ask model-generated questions than template-based.

\subsection{Qualitative Examples}
We conduct qualitative analysis to further investigate how asking questions helps the agent in accomplishing tasks. In summary, in some cases, we observe that while the baseline E.T. model may struggle to predict the correct object, \modelname successfully manages to navigate and manipulate the correct objects by asking relevant questions. 

\begin{figure}[!tb]
\centering
\includegraphics[width=0.99\columnwidth]{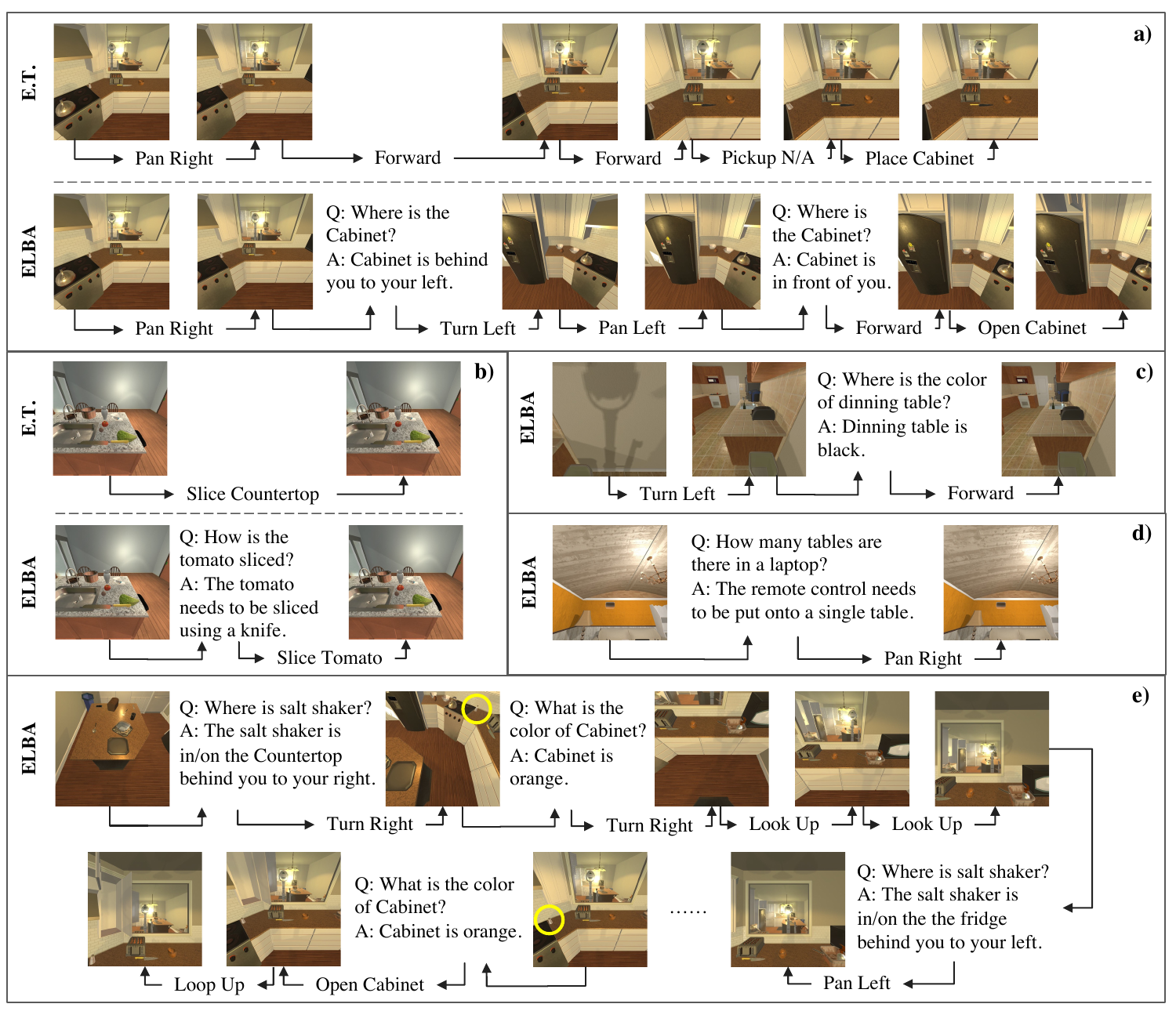}
\vspace{-0.4cm}
\caption{\textbf{Qualitative Examples.} Predicted trajectories of E.T. and \modelname. In each example, the top row shows the predicted trajectory by E.T., and the bottom row shows the predicted trajectory of \modelname. Examples (a) and (b) show successful cases of \modelname, while (c), (d), and (e) show failure cases. Best viewed in color.}
\label{fig:qualitative}
\vspace{-0.3cm}
\end{figure}

We demonstrate two successful examples in Figure~\ref{fig:qualitative} (a) and (b). In each example, the top row shows the predicted trajectory by the E.T. model, and the bottom row depicts the predicted \modelname trajectory. In Figure~\ref{fig:qualitative} (a), we observe that by asking questions about the position of the \texttt{Cabinet}, the agent can find and successfully interact with this object. In Figure~\ref{fig:qualitative} (b), \modelname helps the agent to interact with the correct object (\ie, \texttt{Tomato}) by asking questions, while the E.T. model tries to act on a wrong object (\ie, \texttt{Slice Countertop}). Additional qualitative examples can be found in the Appendix.

We also present a couple of failure cases. In particular, in Figure~\ref{fig:qualitative} (c), \modelname falsely predicts a table's color as ``black'' instead of ``white'', leading the agent to approach the black object in the scene. In Figure~\ref{fig:qualitative} (d), we find that the generated question-answer pair is not well-formed and could not provide helpful information to guide the agent. 
Finally, we also report a case of ill-timed QA in Figure~\ref{fig:qualitative} (e), where the agent goes back and forth asking questions about \texttt{Salt Shaker} and \texttt{Cabinet}. The yellow circle in the figure shows the position of the \texttt{Salt Shaker}. In this example, the agent's goal is to ``\texttt{put all salt shakers in one cabinet}'', where the agent will first need to find the \texttt{Salt Shaker} and then the \texttt{Cabinet}. However, the agent struggles to finish the first sub-goal, which is picking up the \texttt{Salt Shaker}, because it asks an ill-timed question about the next sub-goal (\ie, \texttt{Q: What is the color of Cabinet?}). 

 \section{Conclusion}
To effectively operate in human spaces, autonomous embodied agents need to not only understand and execute instructions but also actively seek supervision to resolve ambiguities naturally arising in real-world tasks. In this work, we introduce \longmodelname (\modelname), an agent that learns when and what to ask for embodied vision-and-language navigation and task completion. Experimental results demonstrate that asking questions leads to improved task performance, opening new directions in task-based interactive embodied QA.

 \section*{Acknowledgments}
\noindent This research is based upon work partially supported by the Amazon - Virginia Tech Initiative for Efficient and Robust Machine Learning and U.S. DARPA ECOLE
Program No. HR00112390062.  The views and conclusions contained herein are those of the authors and should not be interpreted as necessarily representing the official policies, either expressed or implied, of Amazon, DARPA, or the U.S. Government. The U.S. Government is authorized to reproduce and distribute reprints for governmental purposes notwithstanding any copyright annotation therein.

{\small
\bibliographystyle{ieee_fullname}
\bibliography{egbib}
}

\clearpage
\appendix
 
\section{Implementation Details}
\label{sec:imp}
We use the pre-trained weights from the original TEACh codebase for the \textsc{Actioner} and select the confusion thresholds via the TEACh validation set. For clarity and for keeping hyper-parameters minimal, we use the same threshold across both action and object distributions when using entropy-based confusion. Our ablation studies show that using a common hyper-parameter does not substantially affect performance. The confusion threshold is set to $0.9$ for the entropy-based method and $1.2$ for the gradient-based method. Moreover, we train the \textsc{QA Evaluator} on the question-answer pairs extracted from TEACh and the oracle question-answer pairs generated using the \textsc{QA Generator}.
For \textsc{Planner}, we finetune the pre-trained T5 model \cite{raffel2019exploring} using Adam optimizer with the learning rate of $3e-5$ and batch size of $6$. We construct the training data for \textsc{Planner} by converting the training trajectories of TEACH into sequences of subgoals. We treat all interaction actions as subgoals. For navigation actions, we create subgoals by replacing sequences of navigation actions with an abstract ``Find'' action with the destination as the next \texttt{object} manipulated.
We evaluate the performance of \textsc{Planner} via Rouge-L \cite{lin2004rouge}, which measures the longest common subsequence (LCS) between the ground truth sub-goal sequence and the generated sub-goal sequence. For the \textsc{QA Evaluator}, we use a global batch size of $32$, AdamW optimizer \cite{loshchilov2018decoupled} with the weight decay of $0.33$ and learning rate of $1e-5$.
Our code is based on PyTorch \cite{paszke2017automatic} and Huggingface Transformers \cite{wolf2019huggingface}. We train our models on a machine equipped with two RTX 8000 with 40GBs of memory. 

\section{Method}
\subsection{Pseudocode for Entropy-based Confusion}
We provide the pseudocode for our entropy-based confusion module in Algorithm \ref{alg:entropy_confusion}. For clarity, we simplify the question-answer generation and selection by referring to the combination of the \textsc{QA Generator} and \textsc{QA Evaluator} steps as \textsc{Questioner}. 

\begin{algorithm}[!tb]
\caption{Entropy-based Confusion}
\label{alg:entropy_confusion}
\textbf{Input}: Entropy function $H(\cdot)$; Action distribution threshold $\epsilon_\alpha$ ; Object distribution threshold $\epsilon_o$; Interaction action set $A^I$; State information $s_{t-1}\!=\!(x_{1:t-1}, v_{1:t-1}, \alpha_{1:t-1})$; Selected question and answer pair $(q^*_t, a^*_t)$ at time step $t$.
\begin{algorithmic}[1] 
\State $p^\alpha_t, p^o_t \gets \text{\textsc{Actioner}}(s_{t-1})$ \textcolor{gray}{ {\# Select next action}}
\State $\hat{\alpha}_t = \argmax_{\alpha} {p^\alpha_t}$ 
\State $\hat{o}_t = \argmax_{o} {p^o_t}$ 
\If{$\Big(H(p_t^\alpha) > \epsilon_\alpha\Big)$ or $\Big(\hat{\alpha}_t \in A^I \text{ and } H(p_t^o) > \epsilon_o\Big)$}
\State \textcolor{gray}{ {\# Generate question-answer pair}}
    \State $(q^*_t, a^*_t) \gets \text{\textsc{Questioner}}(s_{t-1})$ 
    \State \textcolor{gray}{ {\# Augment state information}}
    \State $\tilde{s}_{t-1} \gets (s_{t-1}, q^*_t, a^*_t)$
    \State \textcolor{gray}{ {\# Select next action given question-answer pair}}
    \State $({\tilde{p}^\alpha_t}, {\tilde{p}^o_t}) \gets \text{\textsc{Actioner}}(\tilde{s}_{t-1})$
    \State \textcolor{gray}{ {\# Compute action and object entropy difference}}
    \State $\Delta_\alpha \gets H({\tilde{p}_t^\alpha}) - H(p_t^\alpha)$
    \State $\Delta_o \gets H({\tilde{p}_t^o}) - H(p_t^o)$
    \If{$\Big(\Delta_\alpha < 0\Big)$ or $\Big(\hat{\alpha}_t \in A^I \text{ and } \Delta_o < 0 \Big)$}
        \State \textcolor{gray}{ {\# If entropy decreases, ask the question}}
        \State $\pi_{\theta}(\tilde{s}_{t-1}) \,{=}\, ({\tilde{p}^\alpha_t}, {\tilde{p}^o_t})$
        \State $\hat{\alpha}_t = \argmax_{\alpha} {\tilde{p}^\alpha_t}$ 
        \State $\hat{o}_t = \argmax_{o} {\tilde{p}^o_t}$ 
    \EndIf
\EndIf
\end{algorithmic}
\end{algorithm}

\subsection{\textsc{QA Evaluator}}

\subsection{Sub-goal Generator}
We further evaluate the sub-goal generator on the seen and unseen test sets employing ROUGE-L and BERTScore as our evaluation metrics. For the immediate next subgoal, ROUGE-L is $66.1$ (seen) and $64.3$ (unseen). When considering the entire sequence of all forthcoming subgoals, the scores were $46.2$ (seen) and $44.1$ (unseen). 
ROUGE-L measures the maximum exact matching subsequence between generated and reference sentences and is considerably high given that our generator produces free-form text. Additionally, utilizing BERTScore, which assesses cosine similarity between contextual embeddings, we observe high scores of $95.2/91.5$ (seen) and $95.0/91.0$ (unseen) for the next subgoal and all subgoals, respectively. This indicates a robust performance in capturing semantic similarity. Manual inspection further corroborated the quality of the generated subgoals, affirming their coherence and logical soundness.

\subsection{Generated QA Pairs}
\label{sec:qa_quality}
To assess the quality of the generated question-answer (QA) pairs, we measure perplexity on the TEACh test split. The generated QA pairs exhibit a lower perplexity of $137.62$, in contrast to the higher perplexity of $316.59$ observed in human-generated QA pairs. This decrease in perplexity indicates an enhanced generalization performance in the generated QA pairs. The higher perplexity in human QA pairs is likely a result of the presence of typos and abbreviations commonly encountered in online text conversations.

Furthermore, we conduct experiments to understand the effect of mismatched QA pairs on the model's efficacy. These experiments involve altering the questions in two specific ways: for the ``Empty Question'' variant, the question is replaced with an empty string, and for the ``$<$UNK$>$ Question'' variant, it is substituted with `$<$UNK$>$'. 
The results, detailed in Table \ref{table:mismatch_qa}, reveal a noticeable decline in performance when the question is substituted with an empty string or `$<$UNK$>$', underscoring the critical role and importance of valid QA pairs.

\begin{table}[!t]
\centering
\caption{\textbf{Impact of Mismatched QA Pairs}}
\vspace{-0.2cm}
\resizebox{0.9\columnwidth}{!}{%
\begin{tabular}{l c c }
\toprule
\textbf{Model} & \textbf{SR [TLW]} & \textbf{GC [TLW]} \\
\midrule
\modelname w/E - Oracle QA & 16.0 [1.5] & 19.4 [4.4] \\ \midrule
- Empty Question & 14.4 [2.4] & 17.6 [4.6] \\ 
- $<$UNK$>$ Question & 13.4 [1.5] & 16.8 [4.0]  \\
\bottomrule
\end{tabular}
}
\label{table:mismatch_qa}
\end{table}

\section{Assessing QA Relevance}\label{sup:qual}
\begin{figure}[t!]
\centering
\includegraphics[width=\columnwidth]{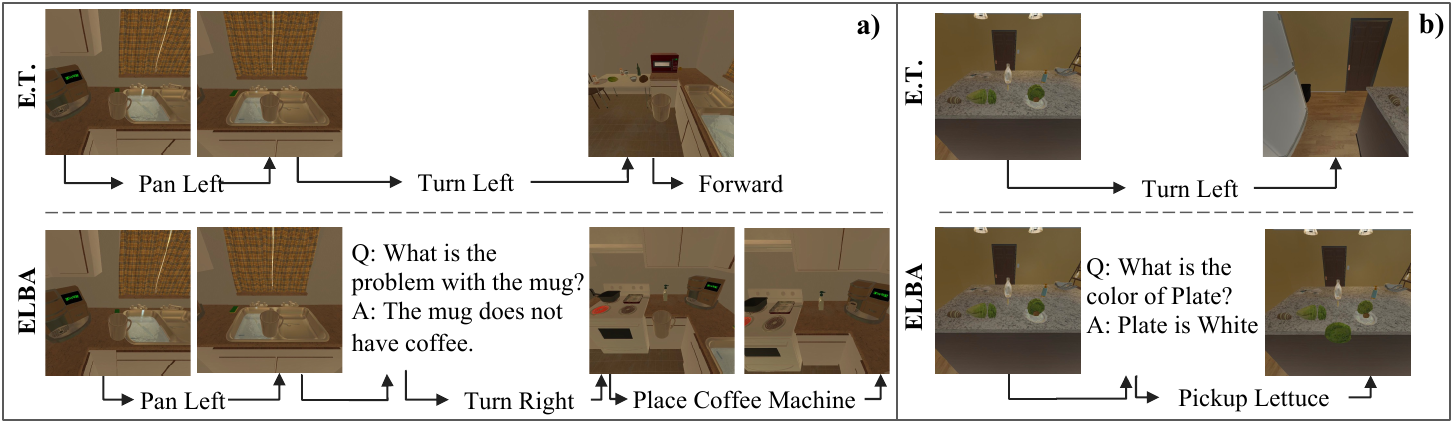}
\vspace{-0.6cm}
\caption{\textbf{Qualitative Examples.}  The predicted trajectory of E.T. and \modelname. In each example, the top row shows the predicted trajectory by the E.T. model, and the bottom row shows the predicted trajectory of \modelname. Examples (a) \texttt{Make coffee} and (b) \texttt{Make breakfast} show successful cases of \modelname.}
\label{fig:add_qualitative}
\end{figure}

Due to the lack of ground truth in both subgoal actions and QAs, assessing the appropriateness of timing and relevance of questions generated by the agent along the trajectories can be challenging. The ideal evaluation would involve a human expert evaluating each question and answer generated across the agent's trajectory, leading to infeasible labor demands.  
Therefore, we instead resort to qualitative analysis, with a few examples shown in Figure \ref{fig:add_qualitative}, and a small-scale user study to evaluate the relevance and correctness of the generated questions for 6 different subgoal tasks.

Figure \ref{fig:add_qualitative} showcases \modelname’s ability to generate QA pairs related to objects critical to the task at hand, thereby guiding the embodied agent to perform actions that are relevant to successfully completing the task. For instance, by querying information about the mug or the color of a plate, the model demonstrates an understanding of the task context required to determine subsequent actions, such as placing the mug in a coffee machine or transferring lettuce to the plate. In contrast, the baseline struggles to discern the most relevant actions and resorts to an exploration of the room. 

\begin{figure*}[t]
\centering
\includegraphics[width=0.8\linewidth]{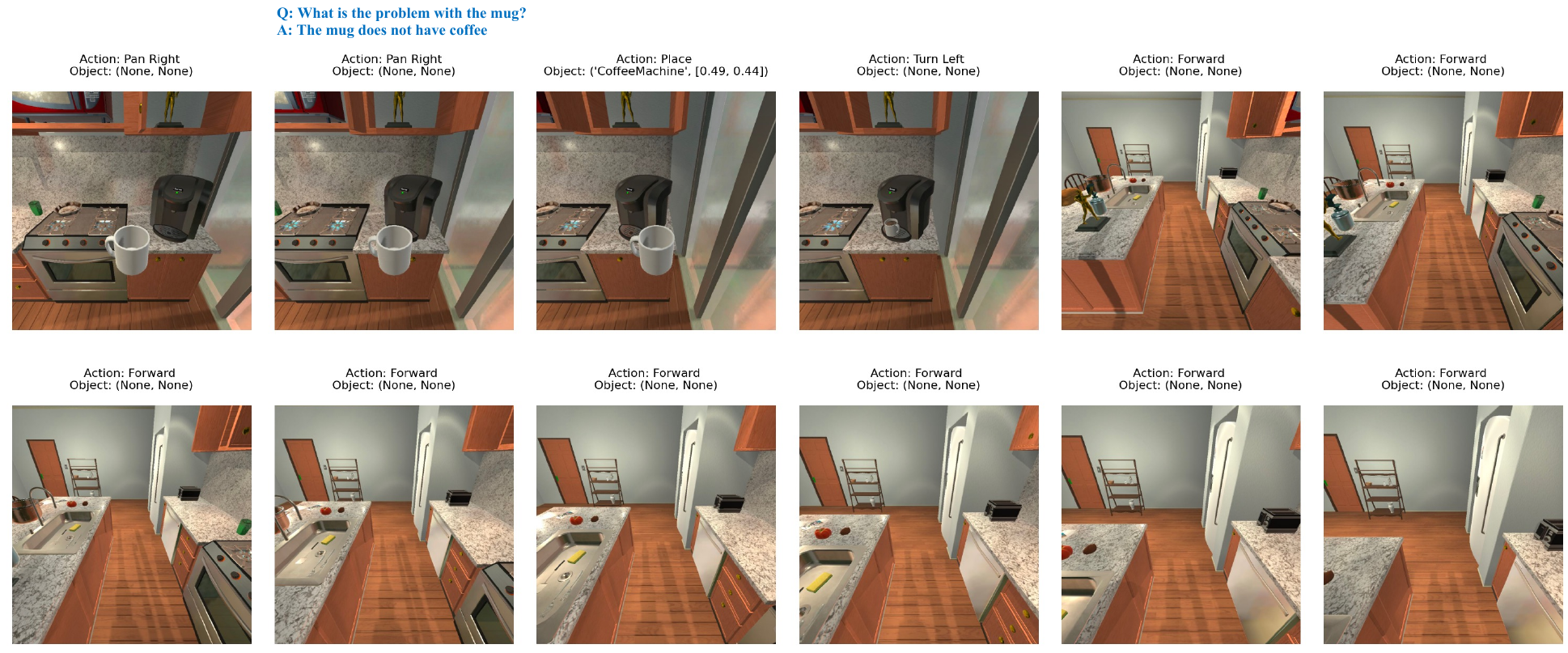}
\vspace{-0.2cm}
\caption{\textbf{User Study Example 1.} Task: \texttt{Coffee}. Goal Condition: \texttt{Place the mug on the coffee machine.} 
}
\label{fig:usr_case1}
\end{figure*}

\begin{figure*}[t]
\centering
\includegraphics[width=0.8\linewidth]{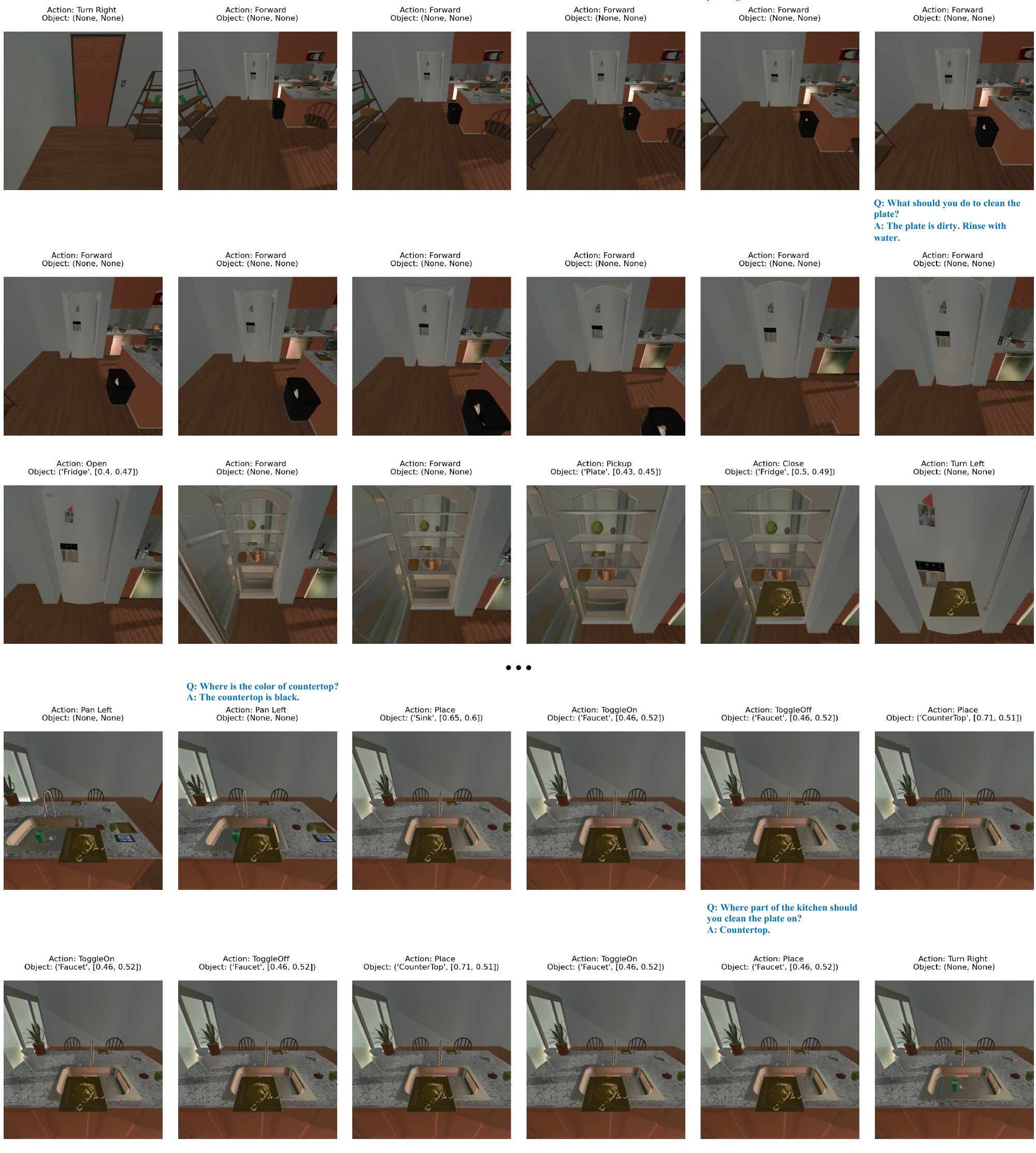}
\vspace{-0.2cm}
\caption{\textbf{User Study Example 2.} Task: \texttt{Clean All X}. Goal Condition:  \texttt{Clean the plate.}
}
\label{fig:usr_case2}
\vspace{-0.5cm}
\end{figure*}

The user study investigates the relevance of the question-and-answer (QA) dialogue in relation to task completion. Participants were presented with six example sub-trajectories depicting \modelname's process of completing various household tasks, with the specific task name and goal condition for each sub-trajectory, and a series of evaluative questions regarding the QA dialogues' relevance to task steps, overall task relevance, grammatical correctness, and any identified issues.
The instructions provided to the participants, as shown in Figure \ref{fig:instructions}, outline the intent of the user study.  Additionally, Figures \ref{fig:usr_case1} and \ref{fig:usr_case2} present two example trajectories featured in the user study and the corresponding task and goal condition presented to the user.

\begin{figure}[htbp]
\centering
\begin{mdframed}[backgroundcolor=gray!20, linewidth=0pt, innerleftmargin=10pt, innerrightmargin=10pt, innertopmargin=10pt, innerbottommargin=10pt]
\footnotesize
\textbf{Instructions:}\\
We present several example trajectories illustrating an agent's process of completing various household tasks. Each example showcases a sequence of images that capture the agent's first-person view of achieving the designated subgoals of the task. On top of each image is text indicating the agent‘s next action. In some of the steps, 
there is an additional question-and-answer dialogue representing the agent’s inquiries at these given time steps during task execution. For each example trajectory, please answer the following questions:

\begin{enumerate}
\item  For each question, is it relevant to the specific time step? If not, please identify the time steps for which a question is irrelevant.
\item  Overall, are the questions posed by the agent relevant to the task?
\item  Are the questions grammatically correct? Please answer `Yes' or `No’.
\item  Do you identify any issues with the questions or answers? Please specify.
\end{enumerate}
\end{mdframed}
\vspace{-0.4cm}
\caption{A snapshot of the user study instructions outlining the objectives and questions.}
\label{fig:instructions}
\end{figure}

\begin{table}[t!]
\centering
\caption{Summarized user study scores - \modelname's QA evaluation.}
\vspace{-0.2cm}
\begin{tabular}{l c }
\toprule
\textbf{Questions} & Percentage \\
\midrule
 Relevance to Task Steps ($\uparrow$) & 61.19\% $\pm$ 15.56 \% \\
 Overall Task Relevance ($\uparrow$) &  80.89\% $\pm$ 18.80\% \\
 Grammatical Correctness ($\uparrow$) & 100\% $\pm$ 0\% \\
 Issues Identified ($\downarrow$) & 62.41\% $\pm$ 18.08\%  \\
\bottomrule
\hline
\end{tabular}
\label{table:user_study_scores}
\end{table}
Table \ref{table:user_study_scores} presents the summarized results of the user study. We compute the percentage of QAs recognized as relevant to the overall task for each instance and average across all examples and participants. This method was similarly applied to the issues flagged by participants.
Participants generally found the QA dialogues relevant to the overarching tasks, with a promising average relevance score of $80.89\%$. However, participants indicated a moderate average score of $61.19\%$ regarding QA relevance to specific task steps, indicating that the question asked might not be directly timely to the next actions to be taken. Despite occasional discrepancies in immediate relevance, the overall task relevance scores show that the \modelname's QA capabilities effectively contribute to task understanding and execution.
All participants confirmed the grammatical correctness of the QA dialogues, underscoring \modelname's ability to generate clear and accurate dialogues. Most of the issues identified are about the repetition of QAs or the relevance of QAs towards specific timesteps. Some users indicated that the question could be relevant to nearby or earlier time steps, suggesting a potential avenue in improving the temporal relevance of QA dialogues during task execution. These findings highlight both strengths and areas of improvement for future research in task-driven interactive QA for embodied agents.

\section{Additional Quantitative Results}
\label{app:helper}
The primary objective of our work is to demonstrate the benefits of enabling an embodied AI agent to ask questions when encountering uncertainty or confusion during task execution. This capability is expected to enhance the performance of an agent by facilitating more effective feedback and decision-making. To validate the general applicability of our approach to different \textsc{ACTIONER} agents, we extend our methodology to HELPER~\cite{sarch2023open}. For this purpose, we integrate a Question-Answering (QA) module within HELPER, that is designed to prompt the agent to ask targeted questions about errors it encounters during task execution, thus providing an opportunity for real-time correction and learning.
In Table \ref{table:helper}, we observe a notable improvement in the performance of HELPER with QA capabilities, suggesting that being able to ask relevant questions can potentially enhance the effectiveness of various \textsc{ACTIONER} models, which are orthogonal contributions to this field.

\section{Additional Qualitative Analysis}

We also analyze the failure cases of \modelname and categorize possible errors into the following limitations:
\vspace{0.1cm}

\noindent \textbf{Color Detection:} The generated oracle QAs sometimes contain errors regarding the appearance of objects. Our model might detect a wrong color, especially when there is a shadow on objects. For example, our model could detect the color of the table as ``black'' while it is supposed to be a ``white'' table under the shadow. Currently, we use a simple dictionary-based approach that first defines a color dictionary that contains the HSV range for each color and then determines the color of an object by looping through the color dictionary and using the color that can cover the largest area as the object color. Thus, there is room for improvement in color detection, \eg, by employing vision models. 
\vspace{0.1cm}

\noindent \textbf{Ill-Formed Model-Generated QAs:} In some cases, the model-generated question-answer pairs might not be well-formed, \eg, when the generated question does not match the candidate answer (\eg, ``\texttt{Q: How is the bowl on the self arranged? A: Place potato in bowl.}'').
This issue could potentially be solved by including an evaluator model that measures the relevance between the question and the answer.

\noindent \textbf{Ill-timed QAs:} We find that the generated question and answer pair at a certain time-step could be ill-timed. For example, when the agent is performing a certain sub-goal (\eg, \texttt{Find Potato}) given a high-level task (\eg, \texttt{Make potato salad}), our model will sometimes generate an ill-timed question on a task-irrelevant sub-goal (\eg, \texttt{Pickup Dish Sponge}) or a sub-goal that follows one or more time steps after the completion of the current sub-goal (\eg, \texttt{Find Plate}). These errors are caused by the fact that we use all future sub-goals predicted by the \textsc{Planner} as candidate answers rather than constructing candidate answers from the next sub-goal instruction only. The latter approach requires the model to track the completion status of the current sub-goal so that the model can decide when to ask questions about the next sub-goal. While our current model bypasses the challenge of tracking sub-goal status by treating all future sub-goals as candidate answers, this leads to ill-timed questions during inference and potentially increases the number of steps needed to complete the task.

\begin{table}[t]
\centering
\caption{Effect of enabling QA in HELPER.}
\vspace{-0.2cm}
\resizebox{0.8\columnwidth}{!}{%
\begin{tabular}{l c c }
\toprule
\textbf{Model} & \textbf{SR [TLW]} & \textbf{GC [TLW]} \\
\midrule
HELPER (reported) & 9.48 [1.21] & 10.05 [3.68] \\
HELPER + QA &  11.05 [1.78] & 13.52 [4.99] \\
\bottomrule
\end{tabular}
}
\label{table:helper}
\vspace{-3mm}
\end{table}

\section{Broader Impact}
Our work highlights the need for a more natural way of interaction for agents to operate in human spaces.
Future extensions of this work include developing more robust QA Evaluators and multimodal QA Generators. While \modelname is a step forward towards truly interactive agents, there remain several open challenges, including but not limited to better contextual understanding and temporal reasoning, handling unexpected or ambiguous feedback, incorporating memory mechanisms to remember and adapt QAs to dynamic changes in the environment during task execution, and automated methods for evaluating timeliness and relevance of task-driven interactive embodied question answering. In future research, we also hope to explore unified generative approaches.

\end{document}